\documentclass[runningheads]{llncs}
\usepackage[T1]{fontenc}
\usepackage{graphicx,verbatim}
\usepackage{booktabs,multirow,graphicx}
\usepackage{makecell}
\usepackage{amssymb}
\usepackage[colorlinks=true, linkcolor=black, citecolor=black, urlcolor=blue]{hyperref}
\usepackage{url}
\usepackage{amsmath} 
\usepackage{cite}
\usepackage{booktabs}
\usepackage{pifont}
\usepackage{booktabs}
\usepackage{pifont}
\usepackage{graphicx} 
\usepackage{booktabs}
\usepackage{pifont}
\usepackage{graphicx} 
\usepackage{makecell} 
\usepackage{multirow}
\usepackage[normalem]{ulem}
\usepackage{makecell}   
\usepackage{multirow}   
\usepackage{booktabs}   
\usepackage{caption}    
\usepackage{graphicx}   
\usepackage{caption}
\usepackage{capt-of} 

\begin{document}
\title{HadBalance: A Plug-and-Play Unified Global Geometric Prior Framework for Generalizable Biomedical Segmentation}
\titlerunning{HadBalance for Generalizable Biomedical Segmentation}
%

\author{
Zhuangzhi Gao\inst{1,2} \and
Feixiang Zhou\inst{1} \and
He Zhao\inst{1} \and
Wenhan Chen\inst{4} \and
Ruiyu Luo\inst{1} \and
Xin Wang\inst{3} \and
Hongyi Qin\inst{5} \and
Zhongli Wu\inst{1,8} \and
Yanda Meng\inst{6} \and
Yitian Zhao\inst{7} \and
Alena Shantsila\inst{8} \and
Gregory Y. H. Lip\inst{8} \and
Eduard Shantsila\inst{2} \and
Yalin Zheng\inst{1}\thanks{Corresponding author.}
}

\authorrunning{Z. Gao et al.}


\institute{
Department of Eye and Vision Science, University of Liverpool, Liverpool, UK
\\
\email{yalin.zheng@liverpool.ac.uk}
\and
Department of Primary Care and Mental Health, University of Liverpool, Liverpool, UK
\and
School of Psychological and Cognitive Sciences, Peking University, Beijing, China
\and
Computer Vision Research Group, University of Amsterdam, Netherlands
\and
Institute of Life Course \& Medical Sciences, University of Liverpool, Liverpool, UK
\and
Bioengineering Program, Biomedical Sciences Division (BioMed), King Abdullah University of Science and Technology (KAUST), Thuwal, Saudi Arabia
\and
Ningbo Cixi Institute
of Biomedical Engineering, Chinese Academy of Sciences, Cixi, China
\and
Liverpool Centre for Cardiovascular Science, University of Liverpool,
Liverpool, UK
}

\maketitle              
\raggedbottom
\begin{abstract} Precise biomedical image segmentation is crucial for clinical diagnosis. Geometric cues (e.g., boundary, shape, and topology) can improve structural consistency, yet most are task-specific and lack a unified geometric foundation that generalizes across organs and modalities. We are motivated by the observation that many medical segmentation targets exhibit globally near-convex shapes. A convex region is one in which any two interior points can be connected by a line segment entirely contained within the region. In practice, medical targets may exhibit small local concavities or boundary irregularities; we refer to such globally convex-like shapes as near-convex. Motivated by this, we derive \textit{Hadwiger Shape Priors} from Hadwiger’s theorem as an interpretable global regularizer using three 2D measures: area $A$, perimeter $P$, and Euler characteristic $\chi$, enabling transfer across organs and modalities. However, because medical datasets are shape-heterogeneous, enforcing near-convex priors uniformly can over-regularize non-convex anatomy (with significant concavities), washing out concavities and fine details and degrading segmentation accuracy. To address this challenge, we propose \textit{Conflict-Aware Objective Balancing (CAOB)}, which integrates shape priors with segmentation in a gradient-aware manner. For each prior, CAOB removes only the gradient component that conflicts with segmentation while preserving the remaining aligned component, and adaptively regulates objective influences to prevent prior dominance. This enables stable use of shape priors on shape-heterogeneous data without erasing genuine concavities or fine structural details. We call this plug-and-play framework \textbf{\textit{HadBalance}}. Our code is available at: \url{https://github.com/NatsuGao7/HadBalance}.


\end{abstract}

\section{Introduction}


Accurate biomedical image segmentation is a cornerstone of clinical diagnosis and has attracted research efforts in recent years, leading to deep learning models tailored to specific organs and imaging modalities that achieve state-of-the-art performance \cite{srivastava2021msrf, shen2017deep,zhou2025glcp, fitzgerald2023fcb, luo2024boundary, gonzalez2025robust}. Incorporating geometric cues (e.g., boundary \cite{yue2022boundary}, shape \cite{wei2025mixture}, and topology \cite{gao2026leveraging}) is a common way to improve structural consistency; however, existing geometric priors are largely task-specific (e.g., topology for vessel connectivity \cite{hu2019topology} and boundary terms for polyp delineation \cite{yue2022boundary}) and \textbf{\textit{lack a unified foundation that generalizes across most organs and modalities and supports transferable geometric constraints}}.

\begin{figure}[t] 
    \centering
    \includegraphics[width=0.8\textwidth]{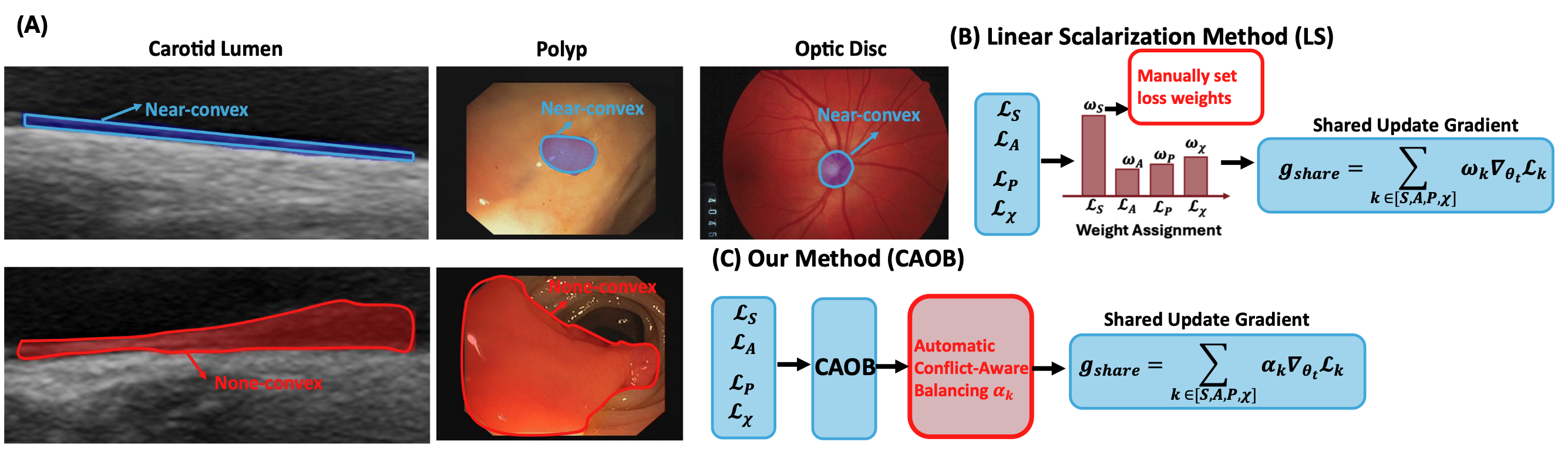} 
    
    \caption{(A)  Examples of near-convex and non-convex anatomical structures.  (B--C) Comparison between LS and CAOB in gradient weighting strategy.
}
    
    \label{fig:motivation} 
\end{figure}

We are motivated by the observation that many medical segmentation targets can be approximated as globally near-convex shapes \cite{liu2020convex, liu2025convex, royer2016convexity}. A convex region is one in which any two interior points can be connected by a line segment entirely contained within the region \cite{boyd2004convex}. As shown in Fig.~\ref{fig:motivation}(A), common anatomical structures—such as polyp regions \cite{bernal2015wm}, the carotid lumen in ultrasound \cite{meiburger2021carotid}, and the optic disc \cite{almazroa2015optic}—are largely convex in their overall shape, despite minor local concavities; we refer to such globally convex-like shapes as near-convex. Motivated by this observation, we draw on Hadwiger's theorem \cite{hadwiger2013vorlesungen}, which provides a complete and minimal basis for shape description via three intrinsic volumes (in 2D: area $A$, perimeter $P$, and Euler characteristic $\chi$; see Section~\ref{sec:prelim}). Accordingly, we propose \textit{Hadwiger Shape Priors} that impose complementary global constraints on the prediction via $A$, $P$, and $\chi$.

However, \textbf{\textit{due to data heterogeneity, near-convex and non-convex anatomies with significant concavities often coexist within the same dataset}} (Fig.~\ref{fig:motivation}(A)). When applied uniformly, Hadwiger Shape Priors may over-regularize non-convex anatomies, suppressing genuine concavities and fine boundary details. A common remedy is to gate priors using heuristic criteria or auxiliary networks; however, reliably distinguishing near-convex from non-convex cases in a unified and generalizable manner is inherently challenging, often introducing additional hyperparameters and unstable decisions on ambiguous samples~\cite{wei2025mixture}. A more principled solution is to address the issue at the optimization level: when shape priors conflict with the segmentation objective, analyzing gradient alignment and adaptively modulating their contributions enables stable and generalizable optimization on heterogeneous data.

Following this idea, a common strategy to resolve gradient conflicts is linear scalarization (LS; Fig.~\ref{fig:motivation}(B)), which combines losses using fixed manual weights \cite{sener2018multi}. However, LS is highly sensitive to weight selection and often requires dataset-specific re-tuning. We instead propose \textit{Conflict-Aware Objective Balancing (CAOB)} (Fig.~\ref{fig:motivation}(C)). At each iteration, CAOB computes gradients of the segmentation loss and the Hadwiger prior terms, treating segmentation as the anchor objective. First, \textit{Primary Gradient Projection (PGP)} removes only the conflicting component of each prior gradient while preserving aligned information. Then, \textit{Adaptive Gradient Balancing (AGB)} solves normalized adaptive coefficients online to regulate the remaining prior contributions, preventing dominance from scale imbalance and ensuring stable integration of geometric priors.

Our key contributions are as follows: \textit{\textbf{1:}} We propose HadBalance, a plug-and-play segmentation framework that unifies geometric shape priors and conflict-aware optimization, achieving strong generalization across diverse organs and modalities. \textit{\textbf{2:}} Motivated by the prevalence of near-convex anatomy in medical data, we introduce Hadwiger-based shape priors as a unified, interpretable regularizer for near-convexity in heterogeneous medical segmentation. \textit{\textbf{3:}} We propose CAOB, which resolves prior–segmentation gradient conflicts without manual weight tuning, enabling stable training on shape-heterogeneous data.

\begin{figure}[t] 
    \centering
    \includegraphics[width=\textwidth]{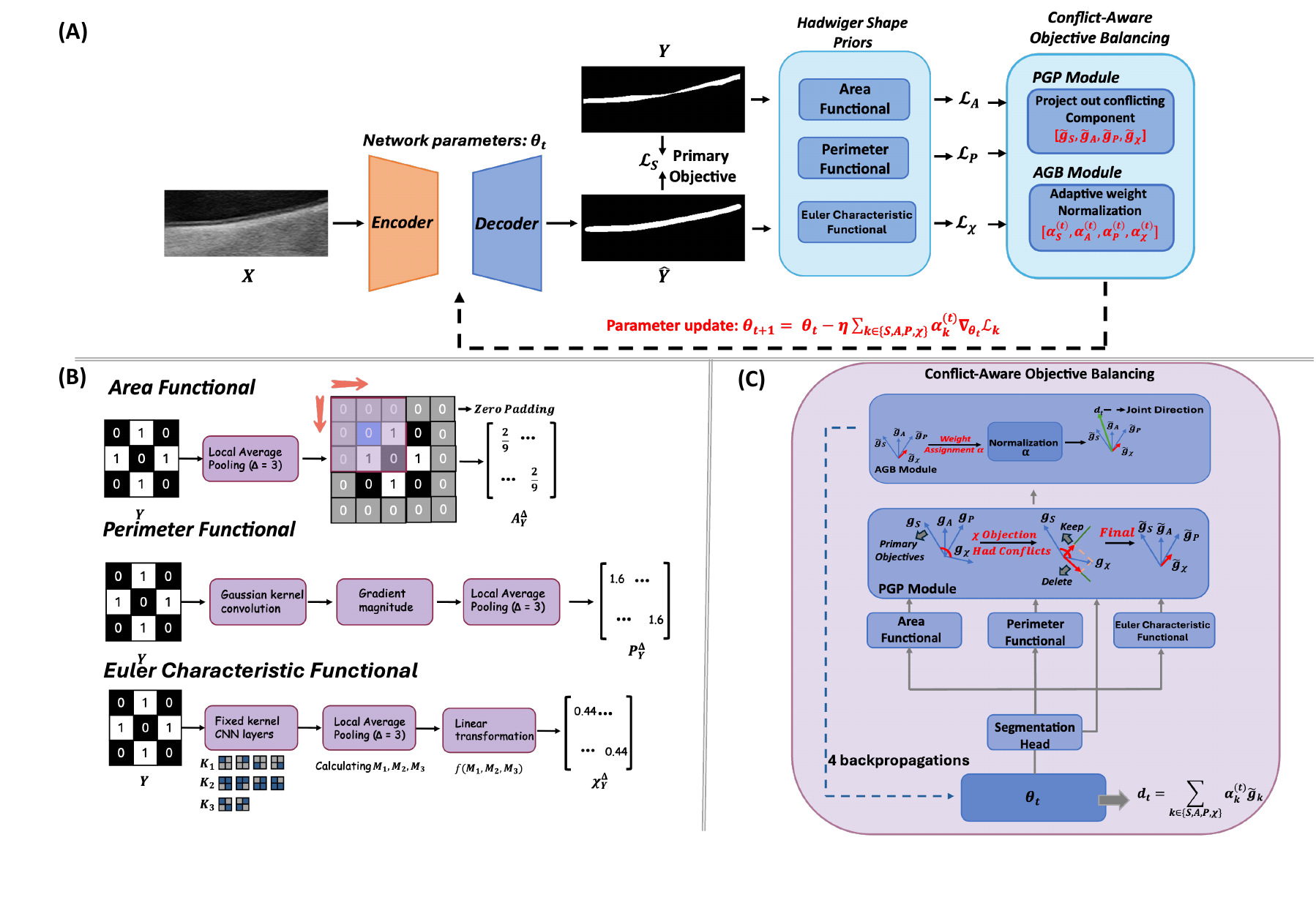} 
    
    \caption{(A) Overview of the framework. 
(B) Computation of the $A$, $P$, and $\chi$ functionals on a ground-truth example. For clarity, we illustrate local average pooling with $\Delta=3$, while $\Delta$ is tunable in practice. (C) CAOB constructs a shared update gradient from per-loss gradients; in the figure, we illustrate the mechanism using a conflict between the $\chi$-prior and the primary gradient.}
    
    \label{fig:overall} 
\end{figure}
\raggedbottom

\section{Methods}\label{sec:prelim}
\noindent \textbf{Preliminaries.} According to Hadwiger's characterization theorem in integral geometry, in $\mathbb{R}^2$, any valuation defined on convex bodies that is motion-invariant, additive, and continuous is a linear combination of the three intrinsic volumes \cite{hadwiger2013vorlesungen}. In the 2D case, these correspond to the $\chi$, $P$, and $A$, respectively. Therefore, any such valuation $\phi(\cdot)$ admits the following representation:
$
\phi(K)=c_0\,\chi(K)+c_1\,P(K)+c_2\,A(K),
\label{eq:hadwiger_2d}
$
where $K \subset \mathbb{R}^2$ denotes a convex body, and $c_0,c_1,c_2$ are constant coefficients. This motivates us to constrain these quantities as principled global shape priors, regularizing the geometric and topological properties of the prediction.

While $A$ and $P$ are standard geometric measures, $\chi$ is a topology-invariant scalar that captures global connectivity beyond pixel-wise accuracy~\cite{hatcher2005algebraic}.
Formally, for a topological space $\Omega$,
$
\chi(\Omega)=\sum_{i=0}^{\infty}(-1)^i \beta_i(\Omega),
$
where $\beta_i$ denotes the $i$-th Betti number.
In practice, we compute $\chi$ on discrete masks by summing local $2\times2$-neighborhood contributions (enumerated by binary configurations, e.g., Gray-code indexing); see Section~\ref{sec:multi_objective} for details~\cite{savage1997survey}.



\raggedbottom
\noindent \textbf{Overall Framework.} As illustrated in Fig.~\ref{fig:overall}(A), given an input image $X$, the segmentation network $f_\theta$ produces a prediction $\hat{Y}=f_\theta(X)$. 
We train the network using a standard segmentation loss $\mathcal{L}_S$ augmented with Hadwiger Shape Priors regularization on $A$, $P$, and $\chi$, implemented as MSE between the corresponding multi-scale local maps of $\hat{Y}$ and the ground truth $Y$.
 In summary, the optimization process can be formulated as 
\begin{equation}
\label{eq:update}
\theta_{t+1} = \theta_t - \eta \sum_{k \in \{S, A, P, \chi\}} \alpha_k^{(t)} \nabla_{\theta_t}\mathcal{L}_k ,
\end{equation}

\noindent where $\theta_t$ denotes the network parameters at iteration $t$, $\eta$ is the learning rate, $\nabla_{\theta_t}\mathcal{L}_k$ denotes the gradient of $\mathcal{L}_k$ with respect to $\theta_t$, and $\alpha_k^{(t)}$ is the weight assigned to loss term $k$ by CAOB module.

\noindent \textbf{Hadwiger Shape Priors.}\label{sec:multi_objective} Incorporating Hadwiger Shape Priors requires differentiable surrogates of the discrete functionals $A, P, \chi$ on soft predictions, and it must account for their heterogeneous units and numerical ranges, which lead to mismatched gradient magnitudes during optimization. Specifically, as illustrated in Fig.~\ref{fig:overall} (B), the area functional computes multi-scale local area maps $A_Y^{\Delta}$ and $A_{\hat{Y}}^{\Delta}$ via $\Delta\times\Delta$ local average pooling (stride $1$) with zero padding, and defines the $A$ prior loss as
$
\mathcal{L}_A=\sum_{\Delta\in\mathcal{D}}\mathrm{MSE}\!\left(A_{\hat{Y}}^{\Delta},A_Y^{\Delta}\right),
$
where $\mathcal{D}$ denotes the set of pooling window sizes used for multi-scale computation. For the perimeter functional, we first obtain a boundary-density map by Gaussian smoothing followed by gradient magnitude (a differentiable proxy of boundary length / total variation), and then apply $\Delta\times\Delta$ average pooling (stride $1$, zero padding) to obtain the local perimeter maps $P_Y^{\Delta}$ and $P_{\hat{Y}}^{\Delta}$. The $P$ prior loss is $
\mathcal{L}_P=\sum_{\Delta\in\mathcal{D}}\mathrm{MSE}\!\left(P_{\hat{Y}}^{\Delta},P_Y^{\Delta}\right).
$ For the Euler characteristic, we adopt the Gray-code (bit-quad) formulation~\cite{savage1997survey}, which decomposes $\chi$ into a sum of local contributions over $2\times2$ binary neighborhoods, to compute multi-scale local $\chi$ maps for $Y$ and $\hat{Y}$. 
We implement this decomposition using fixed-kernel convolutional layers that detect and count the $2\times2$ bit-quad configurations, and then apply $\Delta\times\Delta$ local average pooling (stride $1$, zero padding) to obtain the intermediate maps $M_1^{\Delta}, M_2^{\Delta}, M_3^{\Delta}$. 
The local Euler characteristic map is finally obtained by a linear transformation:
$
\chi^{\Delta}(\cdot)=\frac{M_1^{\Delta}-M_2^{\Delta}-2M_3^{\Delta}}{4}.
$
Accordingly, we obtain $\chi_Y^{\Delta}$ and $\chi_{\hat{Y}}^{\Delta}$ for $Y$ and $\hat{Y}$, and define the $\chi$ prior loss as
$
\mathcal{L}_{\chi}=\sum_{\Delta\in\mathcal{D}}\mathrm{MSE}\!\left(\chi_{\hat{Y}}^{\Delta},\chi_Y^{\Delta}\right).
$ The Hadwiger Shape Priors regularizer is defined as
\begin{equation}
\mathcal{L}_{H}=\mathcal{L}_{A}+\mathcal{L}_{P}+\mathcal{L}_{\chi}.
\end{equation}

\noindent \textbf{Conflict-Aware
Objective Balancing (CAOB).} Unlike LS method that uses a single backward pass on a weighted-sum loss, as illustrated in Fig.~\ref{fig:overall}(C), CAOB requires per-objective gradients; thus, for each mini-batch we compute
$g_k=\nabla_{\theta_t}\mathcal{L}_k$ for $k\in\{S,A,P,\chi\}$ separately.
PGP removes the component of $g_k$ that opposes the segmentation gradient $g_S$ to obtain conflict-free $\tilde g_k$. AGB computes $\{\alpha_k\}$ (via a lightweight non-negative least-squares fit) to balance the $\tilde g_k$ and form the final update direction $d$.

\noindent \textbf{\textit{Primary Gradient Projection (PGP).}}
Using a single shared forward pass, we obtain per-loss gradients $\{g_k\}$ via separate backpropagations. As shown in Fig.~\ref{fig:overall}(C), we treat segmentation as the primary objective. For a prior-term gradient $g_k$, if it is negatively aligned with the segmentation gradient (i.e., $\langle g_k, g_S\rangle<0$), it conflicts with the segmentation objective. We then remove only the component of $g_k$ that opposes $g_S$, yielding a filtered gradient $\tilde g_k$ that satisfies $\langle \tilde g_k, g_S\rangle \ge 0$:
\begin{equation}
\tilde g_k =
\begin{cases}
g_k - \dfrac{\langle g_k, g_S\rangle}{\|g_S\|_2^2+\varepsilon}\, g_S, & \text{if } \langle g_k, g_S\rangle<0,\\[6pt]
g_k, & \text{otherwise,}
\end{cases}
\qquad k\in\{A,P,\chi\},
\end{equation}
with $\tilde g_S = g_S$, where $\varepsilon$ is a constant for numerical stability. This preserves non-conflicting prior information while preventing updates that oppose segmentation.


 
\noindent \textbf{\textit{Alpha-weight Gradient Balancing (AGB).}} PGP resolves directional conflicts, but heterogeneous scales can still let the remaining priors dominate. AGB computes normalized adaptive weights $\{\alpha_k\}$ to form the joint update $d_t$, enabling conflict-aware integration without hand-tuned loss weights. We denote the parameter increment by $\delta\theta_t$ such that $\theta_{t+1}=\theta_t+\delta\theta_t$,
and apply a first-order Taylor expansion, yielding: $\mathcal{L}_k(\theta_{t+1})=\mathcal{L}_k(\theta_t+\delta\theta_t) \approx \mathcal{L}_k(\theta_t)+\nabla_{\theta_t}\mathcal{L}_k(\theta_t)^{\top}\delta\theta_t$. With gradient-based update $\theta_{t+1}=\theta_t-\eta d_t$, we have $\delta\theta_t=-\eta d_t$;
\begin{equation}
\mathcal{L}_k(\theta_{t+1})-\mathcal{L}_k(\theta_t)
\approx -\eta\, \nabla_{\theta_t}\mathcal{L}_k(\theta_t)^{\top} d_t
= -\eta\, \tilde g_k^{\top} d_t .
\end{equation}

We define $u_k^{(t)}=\tilde g_k^{\top} d_t$. If $u_k^{(t)}>0$, $\mathcal{L}_k$ decreases, indicating a beneficial update for objective $k$. The joint update direction is a non-negative combination of the projected gradients, $d_t=\sum_{k}\alpha_k^{(t)}\tilde g_k$. Defining $G_t=[\tilde g_S,\tilde g_A,\tilde g_P,\tilde g_\chi]\in\mathbb{R}^{n\times 4}$ and $\alpha^{(t)}\in\mathbb{R}^4$ gives
$d_t=G_t\alpha^{(t)}$ and $\mathbf u^{(t)}=G_t^{\top}d_t=G_t^{\top}G_t\alpha^{(t)}$. Based on the above, we seek a joint direction $d_t$ that yields positive and balanced utilities across objectives.
Let $K$ denote the number of objectives (here $K=4$). We minimize the average inverse utility
$\min\limits_{d_t}\frac{1}{K}\sum_{k=1}^{K}\frac{1}{u_k^{(t)}}$
which penalizes small utilities more strongly and thus encourages balanced progress.
We further impose the feasibility constraints $u_k^{(t)}\ge 0$ for all $k$ and $\|d_t\|^2\le \rho^2$,
where $\rho$ is a preset constant that controls the overall step scale. We formulate the above as a constrained optimization problem and introduce Lagrange multipliers $\lambda\ge 0$ and $\xi_k\ge 0$ to construct the Lagrangian:
\begin{equation}
\mathcal{J}(d_t,\lambda,\xi)
=\frac{1}{K}\sum_{k=1}^{K}\frac{1}{u_k^{(t)}}
+\lambda\left(\|d_t\|^2-\rho^2\right)
-\sum_{k=1}^{K}\xi_k\,u_k^{(t)},
\qquad \lambda\ge 0,\ \xi_k\ge 0.
\end{equation}
where $\lambda$ and $\xi_k$ are Lagrange multipliers for $\|d_t\|^2\le\rho^2$ and $u_k^{(t)}\ge 0$, respectively. By the Karush--Kuhn--Tucker (KKT) stationarity condition \cite{boyd2004convex}, the optimal direction $d_t^{\ast}$ must satisfy
$\nabla_{d_t}\mathcal{J}(d_t,\lambda,\xi)=0$,
i.e., the first-order derivative of the Lagrangian with respect to $d_t$ vanishes. Taking the derivative w.r.t. $d_t$ and applying complementary slackness (with $u_k^{(t)}>0\Rightarrow \xi_k=0$), we obtain after simplification
$\sum_{k=1}^{K}\frac{1}{\left(u_k^{(t)}\right)^2}\,\tilde g_k
=2K\lambda\, d_t$. Since $2K\lambda$ is a scalar, the above equation indicates that $d_t$ is colinear with
$\sum_{k=1}^{K}\frac{1}{(u_k^{(t)})^2}\tilde g_k$. We thus write
$\mathbf u^{(t)}=G_t^{\top}G_t\alpha^{(t)}$. Together with
$\alpha_k^{(t)}\propto 1/(u_k^{(t)})^2$,
we obtain the fixed-point equation
$
G_t^{\top}G_t\,\alpha^{(t)}=\sqrt{\mathbf 1\oslash \alpha^{(t)}},
$ where the square root is applied element-wise. We solve this fixed-point system for $\alpha^{(t)}$ via a constrained nonlinear least-squares solver (e.g., SciPy \texttt{least\_squares} \cite{virtanen2020scipy}), and then normalize $\alpha^{(t)}$ to remove the inherent scale ambiguity before forming the update direction $d_t=G_t\alpha^{(t)}$.

\section{Experiments and Results}
\textbf{Dataset.} We employed three representative public datasets across different organs: Carotid Ultrasound Boundary (CUBS) \cite{meiburger2021carotid}, Colonoscopy Polyps (CVC-ClinicDB) \cite{bernal2015wm}, and Optic Discs (DRIONS-DB) \cite{Carmona:2008:ION:1383660.1383874}.

\noindent \textbf{Implementation Details.} In our paper, we use multi-scale pooling windows $\mathcal{D}=\{7,15,31\}$. For the LS baseline, all loss weights are set to unity ($w_i=1$) for fair comparison. For CAOB, we warm-start the solver at each iteration with uniform coefficients over the active losses, setting $\alpha_i^{(0)}=1/K$. All experiments are implemented in PyTorch and run on a single NVIDIA RTX A4500 GPU.

\begin{table*}[t]
\centering
\caption{Comparison with representative backbones on CUBS \cite{meiburger2021carotid}, CVC-ClinicDB \cite{bernal2015wm}, and DRIONS-DB \cite{Carmona:2008:ION:1383660.1383874}. The best results are highlighted in \textbf{bold}, while the best among the original backbones are denoted by \underline{\underline{double underlining}}.}
\label{tab:main_results}

\small
\setlength{\tabcolsep}{3pt}
\renewcommand{\arraystretch}{1.15}

\resizebox{\textwidth}{!}{%
\begin{tabular}{l ccc ccc ccc}
\toprule
\multirow{2}{*}{\textbf{Method}} &
\multicolumn{3}{c}{\textbf{CUBS \cite{meiburger2021carotid}}} &
\multicolumn{3}{c}{\textbf{CVC-ClinicDB \cite{bernal2015wm}}} &
\multicolumn{3}{c}{\textbf{DRIONS-DB \cite{Carmona:2008:ION:1383660.1383874}}} \\
\cmidrule(lr){2-4}\cmidrule(lr){5-7}\cmidrule(lr){8-10}
& \textbf{\textit{DSC $\uparrow$}} & \textbf{\textit{IoU $\uparrow$}} & \textbf{\textit{BFScore $\uparrow$}}
& \textbf{\textit{DSC $\uparrow$}} & \textbf{\textit{IoU $\uparrow$}} & \textbf{\textit{BFScore $\uparrow$}}
& \textbf{\textit{DSC $\uparrow$}} & \textbf{\textit{IoU $\uparrow$}} & \textbf{\textit{BFScore $\uparrow$}} \\
\Xhline{1.2pt}

\textbf{UNet} \cite{ronneberger2015u} & 85.56 & 75.12 & 70.26 & 92.12 & 86.51 & 68.56 & 93.95 & 88.64 & 61.08 \\
\textbf{nnUNet} \cite{isensee2021nnu} & 85.23 & 74.52 & 68.23 & \uuline{93.38} & \uuline{88.55} & \uuline{73.46} & \uuline{95.79} & \uuline{92.04} & \textbf{\uuline{71.60}} \\
\textbf{Attention-UNet} \cite{oktay2018attention} & 85.35 & 74.74 & 70.09 & 92.31 & 86.77 & 69.86 & 94.57 & 89.74 & 66.47 \\
\textbf{UNet++} \cite{zhou2018unet++} & 85.30 & 74.68 & 68.73 & 92.04 & 86.37 & 68.54 & 94.23 & 89.14 & 65.13 \\
\textbf{TransUNet} \cite{chen2021transunet} & \uuline{86.02} & \uuline{75.72} & \uuline{70.45} & 91.74 & 86.15 & 66.96 & 94.32 & 89.31 & 66.40 \\
\textbf{SwinUNet} \cite{cao2022swin} & 84.96 & 74.16 & 68.39 & 93.54 & 88.21 & 66.20 & 93.83 & 88.43 & 66.97 \\

\Xhline{1.2pt}
\noalign{\vskip 1.5pt}
\Xhline{1.2pt}

\textbf{+ NASH \cite{navon2022multi}}  & 85.55 & 75.06 & 70.34 & 93.52 & 88.14 & 65.86 & 94.29 & 89.24 & 65.01 \\
\textbf{+ FairGrad \cite{ban2024fair}} & 85.38 & 74.92 & 70.45 & 93.49 & 88.06 & 65.05 & 94.23 & 89.14 & 64.53 \\
\textbf{+ LS}                          & 86.19 & 76.01 & 71.54 & 93.37 & 88.35 & 73.43 & 95.62 & 91.75 & 70.08 \\
\textbf{+ HadBalance}                  & \textbf{86.35} & \textbf{76.24} & \textbf{71.66} & \textbf{94.41} & \textbf{89.85} & \textbf{75.07} & \textbf{95.90} & \textbf{92.25} & 71.18 \\

\bottomrule
\end{tabular}%
}
\end{table*}

\noindent \textbf{Evaluation Metrics and Baselines.} We use the Dice Similarity Coefficient (DSC) and Intersection-over-Union (IoU) to measure overlap between predictions and ground truth, and the Boundary F1-score (BFScore) to evaluate contour accuracy. We evaluate our method on multiple datasets with six representative segmentation backbones, including UNet \cite{ronneberger2015u}, nnUNet \cite{isensee2021nnu}, Attention-UNet \cite{oktay2018attention}, UNet++ \cite{zhou2018unet++}, TransUNet \cite{chen2021transunet}, and SwinUNet \cite{cao2022swin}. For each dataset, we build our method on the strongest-performing backbone. To assess the proposed optimization, we further compare CAOB with Nash-MTL \cite{navon2022multi} and FairGrad \cite{ban2024fair} under identical settings using the same backbone.


\begin{figure}[t] 
    \centering
    \includegraphics[width=0.8\textwidth]{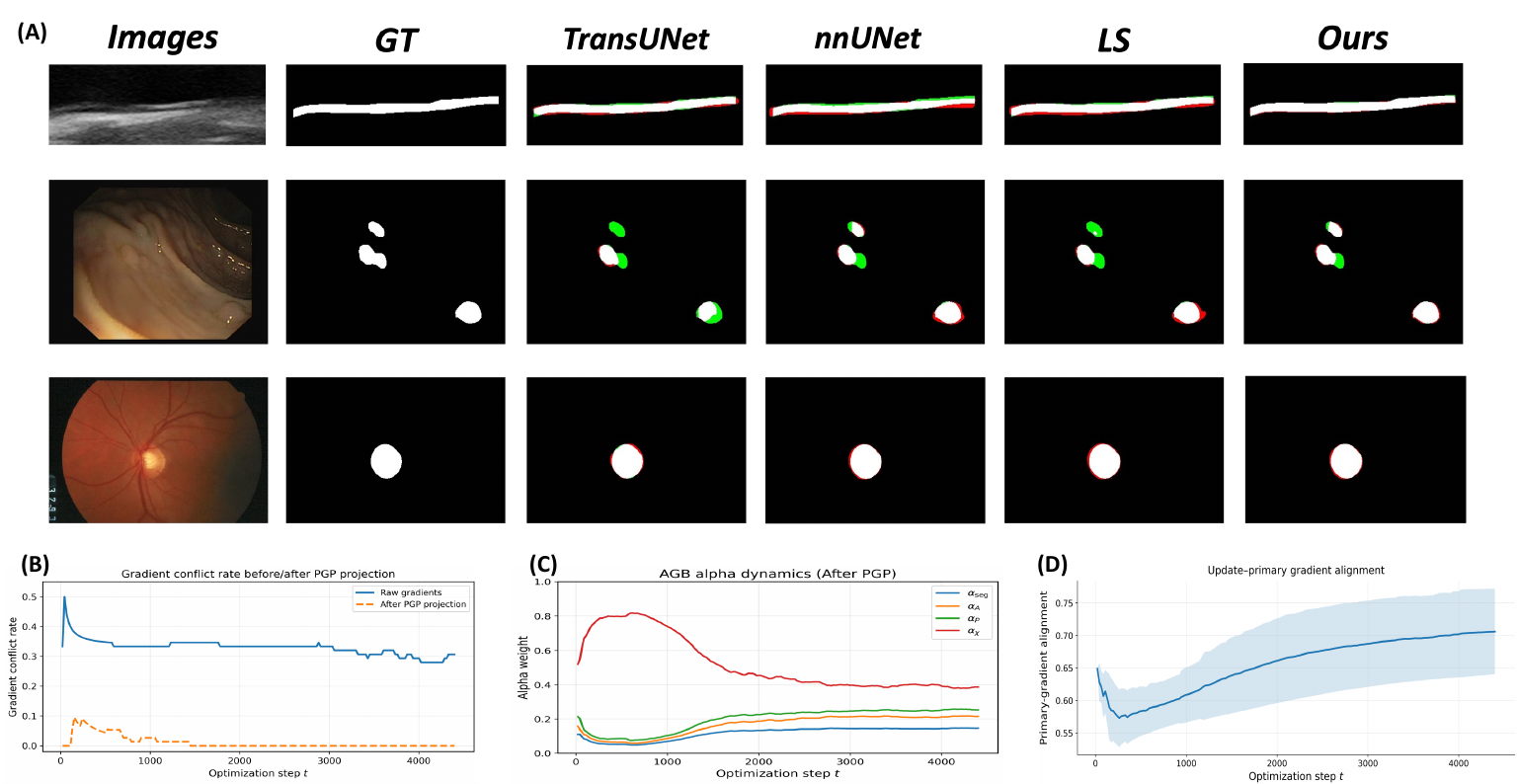} 
    
    \caption{(A) Qualitative results on all datasets with error overlays: white=TP, green=FN, red=FP.
(B) Conflict rate between the segmentation gradient and Hadwiger gradients ($A,P,\chi$): fraction of negative cosine similarities, before and after PGP. 
(C) AGB weight trajectories $\{\alpha_k\}$ after PGP. 
(D) Update--primary alignment (cosine similarity); shaded: rolling mean $\pm$ std.}
    
    \label{fig:figure_4} 
\end{figure}
\raggedbottom
\begin{table}[t]
\newcommand{\cmark}{\ding{51}} 
\newcommand{\xmark}{\ding{55}} 
\centering
\caption{Ablation study of components on CUBS \cite{meiburger2021carotid} and CVC-ClinicDB \cite{bernal2015wm}.}
\label{tab:ablation_multi_dataset}

\small
\renewcommand{\arraystretch}{1.1}
\setlength{\tabcolsep}{3pt}

\resizebox{\linewidth}{!}{%
\begin{tabular}{cccc ccc cccc ccc}
\toprule
\multicolumn{7}{c}{\textbf{CUBS \cite{meiburger2021carotid}}} 
& \multicolumn{7}{c}{\textbf{CVC-ClinicDB \cite{bernal2015wm}}} \\
\cmidrule(lr){1-7}\cmidrule(lr){8-14}

\multicolumn{4}{c}{\textbf{Module / Loss}} & \multicolumn{3}{c}{\textbf{Performance}} 
& \multicolumn{4}{c}{\textbf{Module / Loss}} & \multicolumn{3}{c}{\textbf{Performance}} \\
\cmidrule(lr){1-4}\cmidrule(lr){5-7}\cmidrule(lr){8-11}\cmidrule(lr){12-14}

\textbf{$L_s$} & \makecell{\textbf{Hadwiger}\\\textbf{Priors}} & \textbf{PGP} & \textbf{AGB}
& \textbf{DSC} $\uparrow$ & \textbf{IoU} $\uparrow$ & \textbf{BFScore} $\uparrow$
&
\textbf{$L_s$} & \makecell{\textbf{Hadwiger}\\\textbf{Priors}} & \textbf{PGP} & \textbf{AGB}
& \textbf{DSC} $\uparrow$ & \textbf{IoU} $\uparrow$ & \textbf{BFScore} $\uparrow$ \\
\midrule

\cmark &        &        &        & 86.02 & 75.72 & 70.45
& \cmark &        &        &        & 93.38 & 88.55 & 73.46 \\

\cmark & \cmark &        &        & 86.19 & 76.01 & 71.54
& \cmark & \cmark &        &        & 93.37 & 88.35 & 73.43 \\

\cmark & \cmark & \cmark &        & 86.06 & 75.80 & 71.20
& \cmark & \cmark & \cmark &        & 93.01 & 88.05 & 73.75 \\

\cmark & \cmark &        & \cmark & 85.98 & 75.70 & 70.78
& \cmark & \cmark &        & \cmark & 92.25 & 86.72 & 70.73 \\

\cmark & \cmark & \cmark & \cmark & \textbf{86.35} & \textbf{76.23} & \textbf{71.66}
& \cmark & \cmark & \cmark & \cmark & \textbf{94.41} & \textbf{89.85} & \textbf{75.07} \\

\bottomrule
\end{tabular}%
}
\end{table}
\noindent \textbf{Results.} Table~\ref{tab:main_results} summarizes a comparative evaluation of HadBalance against representative segmentation backbones and specialized gradient-optimization methods on CUBS \cite{meiburger2021carotid}, CVC-ClinicDB \cite{bernal2015wm}, and DRIONS-DB \cite{Carmona:2008:ION:1383660.1383874}. 
On CUBS \cite{meiburger2021carotid}, HadBalance improves the top-performing TransUNet baseline \cite{chen2021transunet} by 0.33 in DSC, 0.52 in IoU, and 1.21 in BFScore, reaching a peak BFScore of 71.66. LS already provides a strong baseline on this near-convex dataset, achieving 86.19 DSC, 76.01 IoU, and 71.54 BFScore; HadBalance yields only marginal additional gains, likely due to limited headroom. On CVC-ClinicDB \cite{bernal2015wm}, which contains more non-convex structures, HadBalance achieves clear improvements, reaching 94.41 DSC, 89.85 IoU, and 75.07 BFScore compared with 93.37 DSC, 88.35 IoU, and 73.43 BFScore from linear scalarization. Without the CAOB modules, linear weighting slightly underperforms nnUNet \cite{isensee2021nnu}, indicating that naive loss weighting is suboptimal in non-convex settings where conflict-aware allocation becomes necessary. In addition, HadBalance outperforms NASH \cite{navon2022multi} and FairGrad \cite{ban2024fair}, with the largest margin on BFScore: 75.07 compared with 65.86 for NASH \cite{navon2022multi} and 65.05 for FairGrad \cite{ban2024fair}, suggesting that generic gradient balancing alone is insufficient to consistently improve boundary quality. As shown in Fig.~\ref{fig:figure_4}(B,C), PGP markedly reduces the conflict rate and stabilizes the subsequent AGB weight dynamics; without PGP, the joint update can be dominated by geometric-prior gradients, particularly the $\chi$ term, steering optimization away from segmentation and degrading boundary quality. On DRIONS-DB \cite{Carmona:2008:ION:1383660.1383874}, HadBalance achieves 95.90 DSC, 92.25 IoU, and 71.18 BFScore, improving over nnUNet by 0.11 in DSC and 0.21 in IoU. Qualitative results are shown in Fig.~\ref{fig:figure_4}(A).

\noindent \textbf{Ablation study.} Table~\ref{tab:ablation_multi_dataset} reports an ablation on CUBS \cite{meiburger2021carotid} and CVC-ClinicDB \cite{bernal2015wm}. Adding Hadwiger Shape Priors improves the BCE baseline on CUBS by 0.17 DSC, 0.29 IoU, and 1.09 BFScore, but slightly reduces performance on CVC-ClinicDB by 0.01 DSC, 0.20 IoU, and 0.03 BFScore, suggesting that Hadwiger priors are sample-dependent and do not transfer without conflict-aware balancing. Using Hadwiger priors as the reference, PGP alone further reduces DSC and IoU on both datasets, indicating that removing negatively aligned prior gradients is insufficient when the remaining priors remain scale-dominant. With PGP and AGB enabled, Fig.~\ref{fig:figure_4}(D) shows increasing alignment between the joint update direction $d$ and the segmentation gradient $g_S$, and performance improves over Hadwiger priors by 0.16 DSC, 0.22 IoU, and 0.12 BFScore on CUBS, and by 1.04 DSC, 1.50 IoU, and 1.64 BFScore on CVC-ClinicDB. Overall, CAOB combines conflict removal and adaptive balancing to enable stable use of Hadwiger priors across datasets containing both near-convex and non-convex targets.

\section{Conclusion}

In this paper, we are motivated by the fact that several medical segmentation targets can be approximated by globally near-convex geometry. To leverage this property, we propose \textbf{\textit{HadBalance}}, which encodes near-convexity via \textit{Hadwiger Shape Priors} ($A$, $P$, and $\chi$) and introduces \textit{CAOB}, a gradient-conflict--aware optimizer that adaptively balances priors against the segmentation objective on shape-heterogeneous data. Experiments on representative benchmarks show consistent improvements. Future work will reduce the additional training cost from extra backward passes.

\section{Disclosure of Interests}
The authors have no competing interests to declare that are relevant to the content of this article.




\bibliographystyle{splncs04}
\bibliography{reference}

@article{meiburger2021carotid,
  title={Carotid ultrasound boundary study (CUBS): An open multicenter analysis of computerized intima--media thickness measurement systems and their clinical impact},
  author={Meiburger, Kristen M and Zahnd, Guillaume and Faita, Francesco and Loizou, Christos P and Carvalho, Catarina and Steinman, David A and Gibello, Lorenzo and Bruno, Rosa Maria and Marzola, Francesco and Clarenbach, Ricarda and others},
  journal={Ultrasound in Medicine \& Biology},
  volume={47},
  number={8},
  pages={2442--2455},
  year={2021},
  publisher={Elsevier}
}

@inproceedings{ronneberger2015u,
  title={U-net: Convolutional networks for biomedical image segmentation},
  author={Ronneberger, Olaf and Fischer, Philipp and Brox, Thomas},
  booktitle={International Conference on Medical image computing and computer-assisted intervention},
  pages={234--241},
  year={2015},
  organization={Springer}
}

@article{bernal2015wm,
  title={WM-DOVA maps for accurate polyp highlighting in colonoscopy: Validation vs. saliency maps from physicians},
  author={Bernal, Jorge and S{\'a}nchez, F Javier and Fern{\'a}ndez-Esparrach, Gloria and Gil, Debora and Rodr{\'\i}guez, Cristina and Vilari{\~n}o, Fernando},
  journal={Computerized medical imaging and graphics},
  volume={43},
  pages={99--111},
  year={2015},
  publisher={Elsevier}
}

@book{hadwiger2013vorlesungen,
  title={Vorlesungen {\"u}ber inhalt, oberfl{\"a}che und isoperimetrie},
  author={Hadwiger, Hugo},
  volume={93},
  year={2013},
  publisher={Springer-Verlag}
}

@article{sener2018multi,
  title={Multi-task learning as multi-objective optimization},
  author={Sener, Ozan and Koltun, Vladlen},
  journal={Advances in neural information processing systems},
  volume={31},
  year={2018}
}

@article{ban2024fair,
  title={Fair resource allocation in multi-task learning},
  author={Ban, Hao and Ji, Kaiyi},
  journal={arXiv preprint arXiv:2402.15638},
  year={2024}
}

@article{navon2022multi,
  title={Multi-task learning as a bargaining game},
  author={Navon, Aviv and Shamsian, Aviv and Achituve, Idan and Maron, Haggai and Kawaguchi, Kenji and Chechik, Gal and Fetaya, Ethan},
  journal={arXiv preprint arXiv:2202.01017},
  year={2022}
}

@book{hatcher2005algebraic,
  title={Algebraic topology},
  author={Hatcher, Allen},
  year={2005},
  publisher={Tsinghua University Press Co., Ltd.}
}

@article{savage1997survey,
  title={A survey of combinatorial Gray codes},
  author={Savage, Carla},
  journal={SIAM review},
  volume={39},
  number={4},
  pages={605--629},
  year={1997},
  publisher={SIAM}
}

@article{fitzgerald2023fcb,
  title={FCB-SwinV2 transformer for polyp segmentation},
  author={Fitzgerald, Kerr and Matuszewski, Bogdan},
  journal={arXiv preprint arXiv:2302.01027},
  year={2023}
}

@article{luo2024boundary,
  title={Boundary fusion multi-scale enhanced network for gland segmentation in colon histology images},
  author={Luo, YuBing and Qin, Pinle and Chai, Rui and Zhai, Shuangjiao and Yan, JunYi},
  journal={Biomedical Signal Processing and Control},
  volume={88},
  pages={105566},
  year={2024},
  publisher={Elsevier}
}

@article{gonzalez2025robust,
  title={Robust t-loss for medical image segmentation},
  author={Gonzalez-Jimenez, Alvaro and Lionetti, Simone and Gottfrois, Philippe and Gr{\"o}ger, Fabian and Navarini, Alexander and Pouly, Marc},
  journal={Medical Image Analysis},
  pages={103735},
  year={2025},
  publisher={Elsevier}
}

@article{isensee2021nnu,
  title={nnU-Net: a self-configuring method for deep learning-based biomedical image segmentation},
  author={Isensee, Fabian and Jaeger, Paul F and Kohl, Simon AA and Petersen, Jens and Maier-Hein, Klaus H},
  journal={Nature methods},
  volume={18},
  number={2},
  pages={203--211},
  year={2021},
  publisher={Nature Publishing Group}
}

@article{srivastava2021msrf,
  title={MSRF-Net: A multi-scale residual fusion network for biomedical image segmentation},
  author={Srivastava, Abhishek and Jha, Debesh and Chanda, Sukalpa and Pal, Umapada and Johansen, H{\aa}vard D and Johansen, Dag and Riegler, Michael A and Ali, Sharib and Halvorsen, P{\aa}l},
  journal={IEEE Journal of Biomedical and Health Informatics},
  volume={26},
  number={5},
  pages={2252--2263},
  year={2021},
  publisher={IEEE}
}

@article{shen2017deep,
  title={Deep learning in medical image analysis},
  author={Shen, Dinggang and Wu, Guorong and Suk, Heung-Il},
  journal={Annual review of biomedical engineering},
  volume={19},
  number={1},
  pages={221--248},
  year={2017},
  publisher={Annual Reviews}
}

@article{oktay2018attention,
  title={Attention u-net: Learning where to look for the pancreas},
  author={Oktay, Ozan and Schlemper, Jo and Folgoc, Loic Le and Lee, Matthew and Heinrich, Mattias and Misawa, Kazunari and Mori, Kensaku and McDonagh, Steven and Hammerla, Nils Y and Kainz, Bernhard and others},
  journal={arXiv preprint arXiv:1804.03999},
  year={2018}
}

@inproceedings{zhou2018unet++,
  title={Unet++: A nested u-net architecture for medical image segmentation},
  author={Zhou, Zongwei and Rahman Siddiquee, Md Mahfuzur and Tajbakhsh, Nima and Liang, Jianming},
  booktitle={International workshop on deep learning in medical image analysis},
  pages={3--11},
  year={2018},
  organization={Springer}
}

@article{chen2021transunet,
  title={Transunet: Transformers make strong encoders for medical image segmentation},
  author={Chen, Jieneng and Lu, Yongyi and Yu, Qihang and Luo, Xiangde and Adeli, Ehsan and Wang, Yan and Lu, Le and Yuille, Alan L and Zhou, Yuyin},
  journal={arXiv preprint arXiv:2102.04306},
  year={2021}
}

@inproceedings{cao2022swin,
  title={Swin-unet: Unet-like pure transformer for medical image segmentation},
  author={Cao, Hu and Wang, Yueyue and Chen, Joy and Jiang, Dongsheng and Zhang, Xiaopeng and Tian, Qi and Wang, Manning},
  booktitle={European conference on computer vision},
  pages={205--218},
  year={2022},
  organization={Springer}
}

@article{Carmona:2008:ION:1383660.1383874,
    author = {Carmona, Enrique J. and Rinc\'{o}n, Mariano and Garc\'{\i}    a-Feijo\'{o}, Juli\'{a}n and Mart\'{\i}nez-de-la-Casa, Jos{\'e} M.},
    title = {Identification of the Optic Nerve Head with Genetic    Algorithms},
    journal = {Artif. Intell. Med.},
    issue_date = {July, 2008},
    volume = {43},
    number = {3},
    month = jul,
    year = {2008},
    issn = {0933-3657},
    pages = {243--259},
    numpages = {17},
    url = {http://dx.doi.org/10.1016/j.artmed.2008.04.005},
    doi = {10.1016/j.artmed.2008.04.005},
    acmid = {1383874},
    publisher = {Elsevier Science Publishers Ltd.},
    address = {Essex, UK},
}

@article{almazroa2015optic,
  title={Optic disc and optic cup segmentation methodologies for glaucoma image detection: a survey},
  author={Almazroa, Ahmed and Burman, Ritambhar and Raahemifar, Kaamran and Lakshminarayanan, Vasudevan},
  journal={Journal of ophthalmology},
  volume={2015},
  number={1},
  pages={180972},
  year={2015},
  publisher={Wiley Online Library}
}

@article{virtanen2020scipy,
  title={SciPy 1.0: fundamental algorithms for scientific computing in Python},
  author={Virtanen, Pauli and Gommers, Ralf and Oliphant, Travis E and Haberland, Matt and Reddy, Tyler and Cournapeau, David and Burovski, Evgeni and Peterson, Pearu and Weckesser, Warren and Bright, Jonathan and others},
  journal={Nature methods},
  volume={17},
  number={3},
  pages={261--272},
  year={2020},
  publisher={Nature Publishing Group US New York}
}

@article{hu2019topology,
  title={Topology-preserving deep image segmentation},
  author={Hu, Xiaoling and Li, Fuxin and Samaras, Dimitris and Chen, Chao},
  journal={Advances in neural information processing systems},
  volume={32},
  year={2019}
}

@article{yue2022boundary,
  title={Boundary constraint network with cross layer feature integration for polyp segmentation},
  author={Yue, Guanghui and Han, Wanwan and Jiang, Bin and Zhou, Tianwei and Cong, Runmin and Wang, Tianfu},
  journal={IEEE Journal of Biomedical and Health Informatics},
  volume={26},
  number={8},
  pages={4090--4099},
  year={2022},
  publisher={IEEE}
}

@inproceedings{wei2025mixture,
  title={Mixture-of-shape-experts (mose): End-to-end shape dictionary framework to prompt sam for generalizable medical segmentation},
  author={Wei, Jia and Zhao, Xiaoqi and Woo, Jonghye and Ouyang, Jinsong and El Fakhri, Georges and Chen, Qingyu and Liu, Xiaofeng},
  booktitle={Proceedings of the Computer Vision and Pattern Recognition Conference},
  pages={6448--6458},
  year={2025}
}

@book{boyd2004convex,
  title={Convex optimization},
  author={Boyd, Stephen and Vandenberghe, Lieven},
  year={2004},
  publisher={Cambridge university press}
}

@inproceedings{zhou2025glcp,
  title={Glcp: Global-to-local connectivity preservation for tubular structure segmentation},
  author={Zhou, Feixiang and Gao, Zhuangzhi and Zhao, He and Xie, Jianyang and Meng, Yanda and Zhao, Yitian and Lip, Gregory YH and Zheng, Yalin},
  booktitle={International Conference on Medical Image Computing and Computer-Assisted Intervention},
  pages={237--247},
  year={2025},
  organization={Springer}
}

@article{gao2026leveraging,
  title={Leveraging Persistence Image to Enhance Robustness and Performance in Curvilinear Structure Segmentation},
  author={Gao, Zhuangzhi and Zhou, Feixiang and Zhao, He and Chen, Xiuju and Li, Xiaoxin and Yu, Qinkai and Zhao, Yitian and Shantsila, Alena and Lip, Gregory YH and Shantsila, Eduard and others},
  journal={arXiv preprint arXiv:2601.18045},
  year={2026}
}

@article{liu2020convex,
  title={Convex shape prior for deep neural convolution network based eye fundus images segmentation},
  author={Liu, Jun and Tai, Xue-Cheng and Luo, Shousheng},
  journal={arXiv preprint arXiv:2005.07476},
  year={2020}
}

@article{liu2025convex,
  title={Convex Shape Prior for Deep Convolution Neural Network-Based Image Segmentation},
  author={Liu, Jun and Zhang, Kehui and Tai, Xue-Cheng and Luo, Shousheng},
  journal={Journal of Mathematical Imaging and Vision},
  volume={67},
  number={6},
  pages={61},
  year={2025},
  publisher={Springer}
}

@inproceedings{royer2016convexity,
  title={Convexity shape constraints for image segmentation},
  author={Royer, Loic A and Richmond, David L and Rother, Carsten and Andres, Bjoern and Kainmueller, Dagmar},
  booktitle={Proceedings of the IEEE Conference on Computer Vision and Pattern Recognition},
  pages={402--410},
  year={2016}
}
\end{document}